# Robust Sequence-to-Sequence Acoustic Modeling with Stepwise Monotonic Attention for Neural TTS

*Mutian He[1*], Yan Deng[2], Lei He[2]*

[1]School of Computer Science and Engineering, Beihang University, China
[2]Microsoft China
mutianhe@buaa.edu.cn, {yaden, helei}@microsoft.com

## Abstract

Neural TTS has demonstrated strong capabilities to generate human-like speech with high quality and naturalness, while its generalization to out-of-domain texts is still a challenging task, with regard to the design of attention-based sequence-to-sequence acoustic modeling. Various errors occur in those inputs with unseen context, including attention collapse, skipping, repeating, etc., which limits the broader applications. In this paper, we propose a novel stepwise monotonic attention method in sequence-to-sequence acoustic modeling to improve the robustness on out-of-domain inputs. The method utilizes the strict monotonic property in TTS with constraints on monotonic hard attention that the alignments between inputs and outputs sequence must be not only monotonic but allowing no skipping on inputs. Soft attention could be used to evade mismatch between training and inference. The experimental results show that the proposed method could achieve significant improvements in robustness on out-of-domain scenarios for phoneme-based models, without any regression on the in-domain naturalness test.

## 1. Introduction

In recent years, approaches of end-to-end neural TTS grow active and have not only exhibited significant advantages in simplicity, but further demonstrated strong capabilities to produce highly natural speech that may rival human recordings. Typical neural TTS models (e.g. Tacotron2) use an attention-based sequence-to-sequence model that takes characters or phonemes as inputs and produces an output sequence of acoustic features (e.g. mel spectrograms), from which raw waveform of high quality could be generated by a neural vocoder, such as WaveNet [1-3].

However, current methods, though satisfying on general inputs similar to its training corpus, are not robust enough when encounter out-of-domain inputs deviated from the training corpus, especially when given long, complicated or abnormal utterances: Unacceptable errors of speech mismatch with input scripts could occur, often with **skipping**, **repeating**, or **attention collapse** (unintelligible gibberish when the model fails to focus on a single input token). Such robustness issue leads to significant barriers in the broader use of neural TTS, when the inputs to the model are more varied and uncontrollable, a resolution of which is hence of great value.

Often such errors could be ascribed to the misalignments from the attention mechanism, which is expected to co-adapt both encoder and decoder to predict the alignments between inputs and outputs [4], that is to say, to determine the corresponding input token to speak out at each output step, and to latently model the durations and pauses in the speech. Therefore, we highlight robustness of attention mechanism as our primary concern, starting from our observation: human read out phones in its original order, each time focused on a short segment of phonemes, and each phone would be read out and contribute to a segment of resultant speech, unlike tasks like speech recognition where continuous input frames (e.g. period of silence) may have no impacts on outputs. Therefore, the alignment should be a *surjective mapping* from output frames to input tokens, following such strict criteria: (1) **Locality**: each output frame is aligned around a single input token, which avoids attention collapse; (2) **Monotonicity**: the position of the aligned input token must never rewind backward, which prevents repeating; (3) **Completeness**: each input token must be once covered or aligned with some output frame, which averts skipping.

Various attention mechanisms have been proposed so as to ensure correct alignments [5-14], while often they would fail to meet all the three criteria. Particularly, completeness in TTS was not thoroughly discussed. In this paper, we investigate attention mechanisms in neural TTS models and propose the novel method of stepwise monotonic attention. Our method, based on monotonic attention, enforces extra constraint for completeness to satisfy all three criteria above. Experiments demonstrate that our method is highly stable even given the most singular inputs, while guarantees naturalness on par with state-of-the-art models on in-domain sentences.

## 2. Related work

### 2.1. Attention-based neural TTS

Generally, acoustic modeling in neural TTS is accomplished by an attention-based encoder-decoder model. Given encoder outputs $x = [x_1, x_2, ..., x_n]$ as "memory" entries, at each timestep $i$, an "energy" value $e_{i,j}$ is evaluated for each $x_j$ by a trainable attention mechanism [1-2, 4], given the previous decoder hidden state $h_{i-1}$, as Equation (1). Then the energy values are normalized to an alignment vector $\alpha_i$ to produce the context vector $c_i$ as Equation (2-3), from which the decoder produces outputs autoregressively.

$$e_{i,j} = Attention(h_{i-1}, x_j) \qquad (1)$$

---

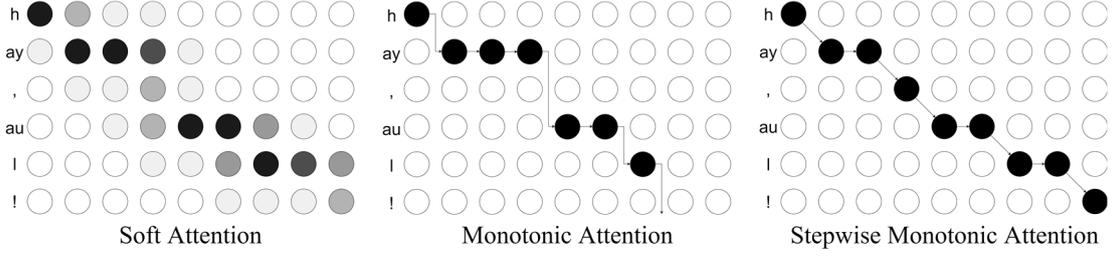

Figure 1: *Illustration for Different Attention Mechanisms*

$$\alpha_{i,j} = \frac{\exp(e_{i,j})}{\sum_{k=1}^{n} \exp(e_{i,k})} = \text{softmax}(e_{i,:})_j \quad (2)$$

$$c_i = \sum_{j=1}^{n} \alpha_{i,j} x_j \quad (3)$$

Recently, numerous variants of attention mechanisms emerge, among which some are specially designed for the property of locality and monotonicity. As for locality, methods like [5-8] are proposed, while monotonicity and completeness could be further improved by explicitly modelling alignments over memories with non-decreasing position [9-12]. Other methods encourage monotonicity and completeness implicitly, by penalizing non-diagonal alignments [13], or utilizing previous alignments as location sensitive attention in Tacotron2 [6]. Particularly, forward attention is proposed to stabilize the attention process in TTS, which reweighs the alignments by previous ones using forward variables, possibly modulated by a "transition agent" [14].

### 2.2. Monotonic attention

Among all the attention variants, monotonic attention has demonstrated to be particularly effective to strictly preserve monotonicity and locality, and has been applied to multiple tasks including TTS [15, 16]. The mechanism could be described as follows [17]: at each step $i$, the mechanism inspects the memory entries from the memory index $t_{i-1}$ it focused on at the previous step. Then energy values $e_{i,j}$ are produced as in Equation (1), but a "selection probability" $p_{i,j}$ is evaluated by logistic sigmoid function instead:

$$p_{i,j} = \sigma(e_{i,j}) \quad (4)$$

Starting from $j = t_{i-1}$, at each time the mechanism would decide to keep $j$ unmoved, or move to next $j \leftarrow j + 1$, by sampling $z_{i,j} \sim \text{Bernoulli}(p_{i,j})$. $j$ would keep moving forward until reaching the end of inputs, or until receiving a positive sampling result $z_{ij} = 1$, and when $j$ stops memory $x_j$ would be directly picked as $c_i$. With such restriction, it is guaranteed that solely one entry would be focused on at each step and its position would never rewind backward. Moreover, the mechanism only requires linear time complexity and supports online inputs, which could be efficient in practice.

Such kind of mechanism with non-differentiable selection of a memory entry as $c_i$ is a "hard" attention, which ensures locality in a radical way, in contrast to the "soft" attention above using continuous weights. Therefore, it could not be trained by standard back-propagation. Multiple approaches have been proposed for this issue, including reinforcement learning [18-20], approximation by beam search [21], and, as in monotonic attention, by soft attention: using the distribution of $p_{i,j}$, which could be recursively evaluated with Equation (5), or in a parallel way with Equation (6) as given in [17]:

$$\alpha_{i,j} = p_{i,j}\left(\frac{(1-p_{i,j-1})\alpha_{i,j-1}}{p_{i,j-1}} + \alpha_{i-1,j}\right) \quad (5)$$

$$\alpha_i = p_i \cdot \text{cumprod}(1 - p_i) \cdot \text{cumsum}\left(\frac{\alpha_{i-1}}{\text{cumprod}(1-p_i)}\right) \quad (6)$$

Then the expectation of the context vector is used instead, in the soft manner same as Equation (3), given $\alpha_{i,j}$. Besides, a few training tricks are applied as in [17] on energy values in Equation (4) before the sigmoid normalization: weight-normalization is applied to mitigate the optimization issues due to the use of sigmoid function in lieu of softmax; a normal-distributed noise is added to encourage close-to-binary $\alpha_{i,j}$; a trainable bias is added to modulate the moving speed.

## 3. Stepwise monotonic attention

As discussed above, monotonic attention meets our criteria of locality and monotonicity, and could be helpful reducing misalignments and improving the robustness of TTS models. To further ensure completeness, we add such restriction: at each decoding step, the hard-aligned position shall move at most one step. We name this mechanism stepwise monotonic attention. In Figure 1 a comparison of attention mechanisms discussed is illustrated.

Similarly, the inference of stepwise monotonic attention involves with sampling: for each timestep $i$, the mechanism inspects the memory entry $j = t_{i-1}$ it attended to at the previous step, and sample from the corresponding Bernoulli distribution as in section 2.2. However, we only need to decide whether move forward by one step or stay unmoved. Therefore, we could directly form the recursive relation of the distribution of $p_{i,j}$ in Equation (7), or more efficiently in Equation (8), where [0; ] stands for zero-padding:

$$\alpha_{ij} = \alpha_{i-1,j-1}(1 - p_{i,j-1}) + \alpha_{i-1,j} p_{ij} \quad (7)$$

$$\alpha_i = \alpha_{i-1} \cdot p_i + [0; \alpha_{i-1,:-1} \cdot (1 - p_{i,:-1})] \quad (8)$$

The expectation of context vectors is used in the same manner at training stage as in Equation (3). Apparently, its linear time complexity and online decoding capability are preserved, and the tricks of noise, trainable bias and normalization on energy values could be applied as well.

We've noticed that our methods share some similar mathematical form with part of forward attention in [14]. However, in our novel work, the specialty of TTS tasks is modeled directly in a simpler and more straightforward way.

An issue could be observed in either the original monotonic attention or the stepwise version that the context vector at the training stage is under different distribution of that at the inference stage. In this paper we refer this issue as *context mismatch*: In training, a "soft" context vector instead of a single memory entry is fed to the decoder, which may lead to issues. For instance, the decoder may learn to predict acoustic features based on the memory entries to be attended in the future other than the current entry that should be attended to. Such mismatch, though might be relieved by close-to-binary $\alpha_{i,j}$ values, or alternative training methods of reinforcement learning, is inherent in such hard attention with soft training stage. Therefore, we propose that we could keep using soft attention at inference as well, with still close-to-binary expectation of the context vector in Equation (3) used as if during training. As a result, at the cost of strict guarantee of locality and monotonicity, as well as its linear time complexity and online decoding capability, the soft inference method could evade the context mismatch and may outperform the hard one under particular cases.

Particularly, it should be noted that in this paper the method is proposed and evaluated based on models for input scripts consisted of phonemes and punctuations. Tokens like punctuations that could not be directly mapped to a segment of speech seemingly break the completeness constraint, but thanks to the highly adaptive neural decoder and bidirectional encoder, the model could capture the true relationship between input tokens and acoustic features that are slightly different from our alignment constraints.

## 4. Experiments

### 4.1. Experiment settings

Tacotron2 models with different attention mechanisms are built up for our experiments, using the same model hyperparameters and acoustic feature settings as in [2] unless other specified. A proprietary English speech corpus with 20.8 hours speech recorded by a professional female speaker was used for training, while input scripts are given in manually checked phonemes instead of characters.

The models on which we conduct experiments include:

1. Baseline: Tacotron2 (with location sensitive attention)
2. GMM attention [10], with 20 mixtures (denoted as *GMM*)
3. Monotonic attention, hard or soft inference (denoted as *MA hard* and *MA soft*)
4. Forward attention, with or without transition agent (denoted as *FA+TA* and *FA w/o TA*)
5. Stepwise monotonic attention, hard or soft inference (denoted as *SMA hard* and *SMA soft*)

Models are distributed on 4 NVIDIA V100 GPUs to train for 120k steps with Adam optimizer to ensure convergence, though all models produce sufficiently clear alignments in 40k steps. Raw waveforms are then synthesized by a WaveNet [3] vocoder trained on the same speech corpus. Both *SMA* and *MA* use initial score bias=3.5 and noise scale=2.0. Particularly, strict gradient clipping must be applied to *FA* models due to frequent gradient explosion, as probabilities are multiplied cumulatively throughout relatively long outputs (given no reduction factor in the Tacotron2 framework, unlike in [14]), which is not the case in *SMA* as energies are gated by $\alpha$. We attempted but failed to train hard attention with reinforcement learning as in [18], possibly due to the inherent high variance.

We evaluate the performance of our models in terms of both naturalness and intelligibility. Two sets are used for naturalness evaluation: 1) General Sentences test set of 80 in-domain sentences held out from training corpus; 2) News test set of 80 out-of-domain sentences with broader contents from news scripts. Intelligibility is evaluated using another test set of 292 long (31.1 words and 129.2 phonemes per case in average), complex, or semantically abnormal utterances highly different from training corpus. Test samples and implementation are available on the author's webpage.[1]

### 4.2. Naturalness evaluation

Crowdsourced preference tests were conducted to compare the naturalness of synthesized speech. All 160 utterances from General Sentence and News set are used, each evaluated by 20 judgers. Results with *p*-values are listed in Table 1-2.

As shown in results, both *MA* models show regression compared to the baseline, especially in News test set with more complicated inputs. Noted issues include skipping words, unnatural prosody, and unstable speed of speech. Therefore, monotonic attention might weaken the performance of TTS models. With soft inference the issue grows more severe, which would be further discussed below. In contrast, both *SMA* models demonstrate naturalness at least on par with the baseline and other peers, which proves that our new method could produce speech without hurt to naturalness generally.

Table 1: *Preference test against Baseline*

| New Model | Baseline Wins | New Model Wins | NP | *p*-value |
|---|---|---|---|---|
| *MA hard* | 31.25% | 23.75% | 45.00% | 0.369 |
|  | **53.75%** | 15.00% | 31.25% | <$10^{-3}$ |
| *MA soft* | 35.00% | 17.50% | 47.50% | 0.030 |
|  | **51.25%** | 7.50% | 31.25% | <$10^{-3}$ |
| *SMA hard* | 21.25% | 32.50% | 46.25% | 0.171 |
|  | 31.25% | 36.25% | 32.50% | 0.589 |
| *SMA soft* | 25.00% | 36.25% | 38.75% | 0.201 |
|  | 31.25% | 40.00% | 28.75% | 0.357 |

Table 2: *Preference test against SMA soft; same settings with Table 1*

| Peer Model | Peer Wins | *SMA soft* Wins | NP | *p*-value |
|---|---|---|---|---|
| *FA+TA* | 45.00% | 35.00% | 20.00% | 0.320 |
|  | 42.50% | 32.50% | 25.00% | 0.305 |
| *GMM* | 16.25% | **63.75%** | 20.00% | <$10^{-3}$ |
|  | 21.25% | **57.50%** | 21.25% | <$10^{-3}$ |
| *SMA hard* | 20.00% | 21.25% | 58.75% | 0.863 |
|  | 15.00% | 18.75% | 66.25% | 0.567 |

**Hint.** NP: No Preference; *p*-value: to reject null hypothesis of no significant preference; For each pair of models, the first line covers results on General Sentences, while the second line on News; Significant results are marked in bold.

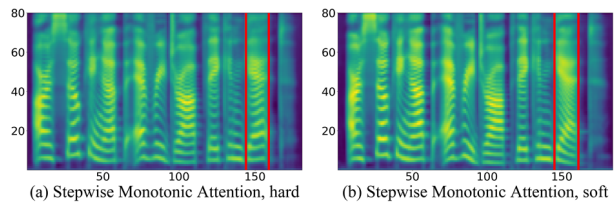

Figure 2: *Mel-spectrogram samples given a plain narrative sentence. Audio available on the webpage.*

---
[1] https://dy-octa.github.io/interspeech2019/index.html

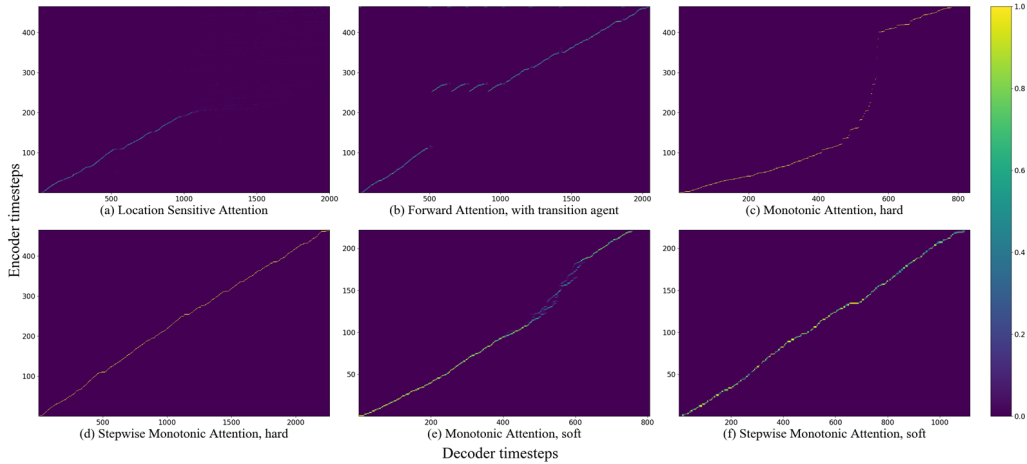

Figure 4: *Alignment samples given out-of-domain inputs, (a)-(d) and (e)-(f) are produced given same inputs, respectively.*

Moreover, though significant differences are not found in the comparison between *SMA hard* and *soft*, prosody issues presumably due to context mismatch could be observed on multiple samples from both *MA* and *SMA hard*, for example as in Figure 2 where we could note an incorrect rising tone within the red marked interval in (a).

### 4.3 Intelligibility evaluation

Intelligibility tests are performed with metrics of case level unintelligible rate (i.e. proportion of cases with significant intelligibility errors) and number of word errors to evaluate the robustness of models given highly singular inputs. As results presented in Figure 3, baseline and *MA* frequently fail to produce intelligible output given out-of-domain texts, while *GMM* models have slightly better results. *FA* methods archive improvements in robustness as claimed in [14]. While *SMA* models have the best robustness in most cases, with only few minor misalignments.

Robustness of different attention mechanisms are further demonstrated in Figure 4 and Table 3. Attention collapse frequently occurs on *Baseline*. As for *MA hard* monotonicity and locality are ensured, which generally avoids attention collapse and repeating, while completeness is frequently violated and a large portion of speech could be skipped. However, the model might limit its errors in a local area but re-stabilize and continue to produce intelligible outputs later. Therefore, longer segments of speech are preserved intelligible in each case, though such refinements are not directly shown on case level results. In both *FA* methods, though with fair improvements, skipping or repeating are still frequent. Among all the models, *SMA* methods demonstrate the best performance, with near-to-perfect correctness in most cases, from which the outstanding robustness could be proved.

Furthermore, though both *MA soft* and *SMA soft* evade context mismatch, neither monotonicity nor locality is strictly guaranteed, which puts robustness at risk. Such unsatisfying scenario could be clearly observed in Figure 4 (e): given out-of-domain inputs, the model may not have sufficient confidence to determine which entry to select, and gradually fall into attention collapse. With context vectors derived from an average of various possible hard alignments with comparable probabilities, completely unintelligible speech is produced. Hence, soft inference in *MA* hurts the results.

However, it is not the case of *SMA*: as shown in Figure 4 (f), given the same out-of-domain inputs of (e), the alignments keep highly discrete: generally, only a single memory entry is focused at each time step and a single alignment path with high likelihood could be determined. We believe that the strong restrictions force the model to learn the alignments with high confidence and locality. As a result, with stepwise monotonic attention, the context mismatch issue could be evaded by soft inference without hurt to its robustness.

Table 3: *Number of Word Errors from Selected Models on Intelligibility Test (Total 9047 words)*

| Model | Attention Collapse | Repeating | Skipping |
|---|---|---|---|
| *Baseline* | 2207 | 79 | 22 |
| *MA hard* | – | 10 | 2660 |
| *FA w/o TA* | 90 | 218 | 493 |
| *SMA soft* | 63 | 37 | 10 |

## 5. Conclusion

This paper proposes the method stepwise monotonic attention. Experiments demonstrate that the method is highly stable and would greatly improve robustness of attention-based neural TTS models under out-of-domain scenarios. The method could be further applied to other sequence-to-sequence tasks similar to TTS, including audio, handwriting and inflection generation. Future work would focus on its refinement on TTS tasks, especially on the character-based scenarios which are further deviated from our alignment constraints, and exploring its application on other fields.

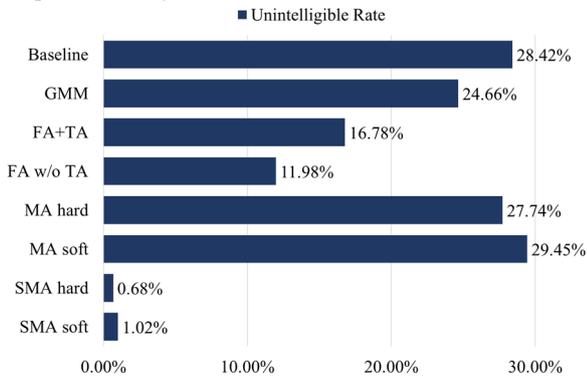

Figure 3 *Intelligibility results (case level)*

Unintelligible Rate:
- Baseline: 28.42%
- GMM: 24.66%
- FA+TA: 16.78%
- FA w/o TA: 11.98%
- MA hard: 27.74%
- MA soft: 29.45%
- SMA hard: 0.68%
- SMA soft: 1.02%

# 6. References

[1] Y. Wang, R. J. Skerry-Ryan, D. Stanton, *et al.*, "Tacotron: Towards End-to-End Speech Synthesis," in *Proceedings of Interspeech 2017*, pp. 4006-4010, 2017.

[2] J. Shen, R. Pang, R. J. Weiss, *et al.*, "Natural TTS Synthesis by Conditioning Wavenet on Mel Spectrogram Predictions," in *2018 IEEE International Conference on Acoustics, Speech and Signal Processing (ICASSP)*, pp. 4779–4783, 2018.

[3] A. van den Oord, S. Dieleman, H. Zen, *et al.*, "WaveNet: A Generative Model for Raw Audio," *arXiv preprint arXiv:1609.03499*, 2016.

[4] D. Bahdanau, K. Cho and Y. Bengio, "Neural Machine Translation by Jointly Learning to Align and Translate," *arXiv preprint arXiv:1409.0473*, 2014.

[5] T. Luong, H. Pham and C. D. Manning, "Effective Approaches to Attention-based Neural Machine Translation," in *Proceedings of the 2015 Conference on Empirical Methods in Natural Language Processing*, pp. 1412-1421, 2015.

[6] J. K. Chorowski, D. Bahdanau, D. Serdyuk, *et al.*, "Attention-based Models for Speech Recognition," in *Advances in Neural Information Processing Systems*, vol. 28, pp. 577-585, 2015.

[7] A. F. T. Martins and R. F. Astudillo, "From softmax to sparsemax: a Sparse Model of Attention and Multi-label Classification," in *Proceedings of the 33rd International Conference on Machine Learning*, pp. 1614-1623, 2016.

[8] V. Niculae and M. Blondel, "A Regularized Framework for Sparse and Structured Neural Attention," in *Advances in Neural Information Processing Systems*, vol. 30, pp. 3338-3348, 2017.

[9] N. Jaitly, D. Sussillo, Q. V. Le, *et al.*, "An Online Sequence-to-Sequence Model Using Partial Conditioning," in *Advances in Neural Information Processing Systems*, vol. 29, 5067–5075, 2016.

[10] A. Graves, "Generating Sequences with Recurrent Neural Networks," *arXiv preprint arXiv:1308.0850*, 2013.

[11] J. Hou, S. Zhang and L. R. Dai, "Gaussian Prediction Based Attention for Online End-to-End Speech Recognition," in *Proceedings of Interspeech 2017*, pp. 3692-3696, 2017.

[12] A. Tjandra, S. Sakti and S. Nakamura, "Local Monotonic Attention Mechanism for End-to-End Speech and Language Processing," in *Proceedings of the 8th International Joint Conference on Natural Language Processing (Volume 1: Long Papers)*, pp. 431–440, 2017.

[13] H. Tachibana, K. Uenoyama and S. Aihara, "Efficiently Trainable Text-to-Speech System Based on Deep Convolutional Networks with Guided Attention," in *2018 IEEE International Conference on Acoustics, Speech and Signal Processing (ICASSP)*, pp. 4784–4788, 2018.

[14] J. X. Zhang, Z. H. Ling and L. R. Dai, "Forward Attention in Sequence-to-Sequence Acoustic Modeling for Speech Synthesis," in *2018 IEEE International Conference on Acoustics, Speech and Signal Processing (ICASSP)*, pp. 4789–4793, 2018.

[15] R. Aharoni and Y. Goldberg, "Morphological Inflection Generation with Hard Monotonic Attention," in *Proceedings of the 55th Annual Meeting of the Association for Computational Linguistics (Volume 1: Long Papers)*, pp. 2004–2015, 2017.

[16] W. Ping, K. Peng, A. Gibiansky, *et al.*, "Deep Voice 3: Scaling Text-to-Speech with Convolutional Sequence Learning," *arXiv preprint arXiv: 1710.07654*, 2017.

[17] C. Raffel, D. Eck, P. Liu, *et al.*, "Online and Linear-Time Attention by Enforcing Monotonic Alignments," in *Proceedings of the 34th International Conference on Machine Learning*, pp. 2837–2846, 2017.

[18] K. Xu, J. Ba, R. Kiros, *et al.*, "Show, Attend and Tell: Neural Image Caption Generation with Visual Attention," in *Proceedings of the 32nd International Conference on Machine Learning*, pp. 2048–2057, 2015.

[19] W. Zaremba and I. Sutskever, "Reinforcement Learning Neural Turing Machines," *arXiv preprint arXiv:1505.00521*, 2015.

[20] J. Ling and A. M. Rush, "Coarse-to-Fine Attention Models for Document Summarization," in *Proceedings of the Workshop on New Frontiers in Summarization*, pp. 33-42, 2017.

[21] S. Shankar, S. Garg and S. Sarawagi, "Surprisingly Easy Hard-Attention for Sequence to Sequence Learning," in *Proceedings of the 2018 Conference on Empirical Methods in Natural Language Processing*, pp. 640-645, 2018.